\documentclass[runningheads]{llncs}
\usepackage[T1]{fontenc}
\usepackage{graphicx}
\usepackage{xcolor}

\usepackage{amsmath}
\usepackage{float}
\usepackage{amssymb}
\usepackage[linesnumbered,ruled,vlined]{algorithm2e}
\usepackage{booktabs}
\SetKwInput{KwInput}{Input}                
\SetKwInput{KwOutput}{Output}              

\definecolor{mmcolor}{rgb}{0.04,0.45,0.15}
\usepackage[normalem]{ulem}

\begin{document}
%
\title{Implicit Shape-Prior for Few-Shot Assisted 3D Segmentation\thanks{This preprint has not undergone peer review or any post-submission improvements or corrections. 
The Version of Record of this contribution will be published in a Springer Nature Computer Science book series 
(CCIS, LNAI, LNBI, LNBIP or LNCS) and the doi will soon be released.}}
%
%
\author{Mathilde Monvoisin\inst{1}\thanks{Both first Authors contributed equally to this work, lastnames in alphabetical order.} \and
Louise Piecuch\inst{1} \and
Blanche Texier \inst{4} \and 
Cédric Hémon \inst{4} \and 
Anaïs Barateau \inst{4}  \and 
Jérémie Huet \inst{2}$^,$\inst{3}\and
Antoine Nordez \inst{2}$^,$\inst{5}\and
Anne-Sophie Boureau \inst{3}\and
Jean-Claude Nunes \inst{4}\and 
Diana Mateus \inst{1}
}
\authorrunning{M. Monvoisin, L. Piecuch et al.}
%
\institute{Nantes Université, École Centrale Nantes, CNRS, LS2N, UMR 6004, 
 \and Nantes Universit\'e, Movement Interactions Performance, IP~UR 4334 UFR STAPS,
\and Nantes Université, CHU Nantes, Pole de Gérontologie Clinique, \\
$^{1,2,3}$ F-44000, Nantes, France
\and Univ Rennes, CLCC Eug\`ene Marquis, INSERM, LTSI - UMR 1099, F-35000 Rennes, France
\and Institut Universitaire de France  (IUF),  Paris,  France\\
\email{mathilde.monvoisin@ls2n.fr}\\
\email{louise.piecuch@ls2n.fr}}
%
\maketitle              
\begin{abstract} 
The objective of this paper is to significantly reduce the manual workload required from medical professionals in complex 3D segmentation tasks that cannot be yet fully automated.
For instance, in radiotherapy planning, organs at risk must be accurately identified in computed tomography (CT) or magnetic resonance imaging (MRI) scans to ensure they are spared from harmful radiation.
Similarly, diagnosing age-related degenerative diseases such as sarcopenia, which involve progressive muscle volume loss and strength, is commonly based on muscular mass measurements often obtained from manual segmentation of medical volumes.
To alleviate the manual-segmentation burden, this paper introduces an implicit shape prior to segment volumes from sparse slice manual annotations generalized to the multi-organ case, along with a simple framework for automatically selecting the most informative slices to guide and minimize the next interactions. The experimental validation shows the method's effectiveness on two medical use cases: assisted segmentation in the context of at risks organs for brain cancer patients, and acceleration of the creation of a new database with unseen muscle shapes for patients with sarcopenia. 

\end{abstract}
\section{Introduction}

Medical image segmentation remains at the heart of many challenges. Whether it is for monitoring cancer patients, in radiotherapy planning, where segmenting organs at risk on Magnetic Resonance (MR) or Computer Tomography (CT) scans  \cite{cubero2025deep} is crucial for better patient care, or in geriatrics or physical and rehabilitation medicine, where the segmentation of certain 
muscles from 3D ultrasound (3DUS) can aid in diagnosis or monitoring of certain diseases such as sarcopenia.
Despite the existence of many fully automated methods, the gold standard in clinical practice remains the manual delineation of the organs by an expert. Most fully automatic methods are based on Convolutional Neural Networks (CNNs) and transformers require a large amount of annotated data, do not automatically generalize across different imaging modalities, and provide pixel-based predictions \cite{unet}\cite{unetr}. As a result, when training on small datasets, the predictions can yield anatomically unrealistic segmentations, with holes or isolated regions \cite{lin2024novel}. 
While foundation models like the Segment Anything Model (SAM)\cite{kirillov2023segment} are widely adopted today, they tend to be computationally heavy, potentially unstable, and their performance often depends on the clarity of object boundaries \cite{huang2024segment} \cite{cheng2023sam}. 
This poses a challenge for organs at risk on CT or sarcopenic muscles in 3DUS images, where contours lack contrast or are noisy. These challenges motivate us to explore interactive segmentations and implicit shape priors.

Recent work on interactive medical image segmentation leverages expert knowledge to improve accuracy while reducing annotation effort. 
Memory-based CNN approaches, such as those by Mikhailov et al \cite{MIKHAILOV2024108038} and Tian et al. \cite{tian2023dynamic}, analyze user corrections over time, supporting effective multi-slice interactive segmentation. MedUHIP \cite{zhu2024meduhip} incorporates uncertainty modeling to generate and refine multiple plausible segmentations, while Cerqueira et al. \cite{cerqueira2024interactive} demonstrate that combining expert-chosen slices with sparse supervision can rival fully supervised methods.
DeepEdit \cite{diaz2022deepedit} and PE-MED \cite{chang2023pe} 
rely on a neural network backbone combined with user clicks and prompt-enhanced feedback to iteratively refine 3D segmentations. SISeg \cite{li2024adaptive}, built on SAM2, actively selects the most informative slices across imaging modalities to guide expert input.  
The above strategies highlight the prevalent role of 
expert-in-the-loop systems.
However, to the best of our knowledge, they do not make use of shape priors and and are still subject to the aforementioned limitations. Therefore, we have chosen to investigate the use of implicit methods within an interactive framework.

Statistical Shape Models (SSMs) have been widely used as explicit shape priors for segmentation. For example, DeepSSM \cite{bhalodia2024deepssm} predicts shape representations from 3D images based on a Point Distribution Model (PDM) built from shapes with predefined correspondences. Likewise, Adams et al. \cite{adams2024weakly}, rely on BVID-DeepSSM to predict anatomically probabilistic shapes from learned correspondences. 
MASSM \cite{ukey2024massm} generalizes shape representations across multiple anatomies enabling multiple shape delineation in image space. 
Despite methodological differences, a bottleneck from above approaches consist in finding point correspondences at some stage. 
Other explicit methods, such as shape-from-template approaches, also enable the creation of multi-organ surface segmentations. 
For instance, Bongratz et al. \cite{bongratz2023abdominal} proposed  UNetFlow, which learns a diffeomorphic deformation field from a reference abdominal CT or MRI and align a template mesh to the predicted voxel segmentation.
However, obtaining such templates is similarly non-trivial. Therefore, we focus on implicit methods, which bypass the need for point correspondences or templates and enable direct shape learning from data.


Regarding implicit shape priors, Amiranashvili et al. \cite{AMIRANASHVILI2024103099} introduced an implicit modeling approach using an auto-decoder 
(AD) trained on sparse binary masks with large inter-slice spacing. Romana De Paolis et al. \cite{de2024fast} proposed a meta-learning implicit method for rapid reconstruction of anatomical shapes from partial data. 
However, both \cite{AMIRANASHVILI2024103099} and \cite{de2024fast} are limited to single-label segmentation. 
Jouvencel et al. \cite{10980745} addressed this challenge 
by developing a multi-label implicit method that incorporates image data, but relies on contour-derived point clouds limiting its adaptability to certain modalities.

This work builds upon the implicit neural representation in \cite{AMIRANASHVILI2024103099}, which models a shape prior across a population. The method was designed to reconstruct full volumetric segmentations from sparsely annotated 2D slices of a single anatomical structure, leveraging the learned prior to facilitate manual segmentation tasks. In this study, we extend their approach to support multi-organ learning and introduce a more efficient slice selection strategy. 
To show the generality of the proposed method, we present results for two practical use cases where the gold standard today is manual or semi-automatic segmentation, namely organ-at-risk segmentation for image guided radiotherapy from CT/MR, and sarcopenia diagnosis from 3DUS scans. For the first application, the method allows clinicians to provide only a single slice per organ to be segmented, and then focus on supplying the slices with the highest prediction error in order to improve the segmentation. The second application aims to accelerate the segmentation of abnormally shaped muscles while taking advantage of a shape prior built from normally shaped muscles. In both cases, an improvement in segmentation quality can be observed, particularly when the number of selected slices is very limited, demonstrating good generalization across multiple anatomies.






\section{Method}
\label{sec:method}


The main goal of our study is to conceive an interactive segmentation method with minimal interactions based on a shape prior. We learn an implicit shape prior for the target organs from a small population of pre-segmented volumes. The learned prior then guides the subsequent expert interactions aiming to reduce their number. 
In practice, the shape model is an implicit coordinate-based neural network trained to predict a probabilistic occupancy for each coordinate. Implicit models are lightweight, provide a continuous representation that can handle variable resolutions and missing data. More precisely we rely on an AD, which expects a few 2D segmentation masks as input, and is therefore image modality agnostic. In this work we focus on a multi-organ extension and on guiding the interactions with a simple slice selection approach based on previous errors.

More formally, given a training dataset of $N$ segmented volumes $\{\mathbf{Y}_i\}_{i=1}^{N}$, a generative decoder $f_\theta$ learns the expected population shape 
during training. 
In our case, the shape $\mathbf{Y}_i$ 
can be a binary or multi-class volumetric occupancy grid. In addition to the network parameters $\theta$, the 
AD optimizes a latent vector $\mathbf{z}_i$ per volume updated both during the training and inference optimization steps. This latent representation will implicitly encode an individual shape prior, enabling the generation of complete 3D shapes from sparse 2D slices. Our method consists of two steps. First, a minimal subset of slices is created.
In a second time, the model's shape prediction is compared with the true shape, and the slices corresponding to the highest errors are selected for the full shape inference. 
The method was applied to two specific use cases as further explained in Sec.~\ref{subsec:selection_slices}.


 

\begin{figure}[htbp]
\centering
\includegraphics[width=1\textwidth]{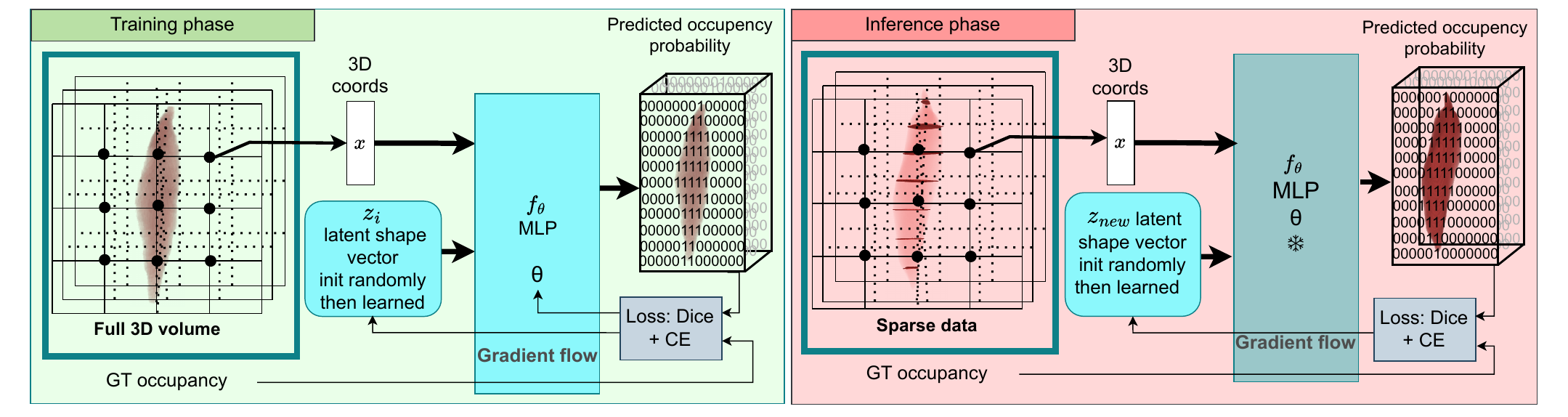}
\caption{Implicit shape decoder. The training phase (\textbf{left}) provides the full segmented volume as target to learn the latent shape vector $z_i$. The inference phase (\textbf{right}) takes only some slices as input to predict a full segmentation with the previously learned $z_i$.}
\label{fig:architecture}
\end{figure}


\subsection{Implicit Prior Shape Learning}
Our implicit model consist of a single Multilayer Perceptron (MLP) used as a per-voxel classifier $f_\theta$, to obtain a binary or multi-class occupancy grid. As shown in Fig. \ref{fig:architecture}, the 3D coordinates $ \mathbf{x} \in \mathbb{R}^3 $ of all points are given one-by-one to the model. The MLP, conditioned by the shape latent vector $ \mathbf{z} \in \mathbb{R}^d$, 
provides a probabilistic prediction for the output voxel as $f_\theta : \mathbb{R}^3 \times \mathbb{R}^d \to [0,1]^{N_{\rm class}}$
where $ N_{\rm class} $ is the number of organs to segment including one for the background. Thereby, the MLP estimates a one-hot prediction $\hat{\mathbf{y}}$ for the voxel at $\mathbf{x}$ conditioned by shape prior $z$, indicating its probability to belong to one of the organs or background, \textit{i.e.} $\hat{\mathbf{y}} = f_\theta(\mathbf{x}, \mathbf{z})$, with $\sum_{c=1}^{N_{\rm class}}\hat y_c =1$ and $\hat y_c$ the c-th element of $\hat{\mathbf{y}}$. 

\subsection{Training Phase}
During training (see Fig.~\ref{fig:architecture}-left), each complete 3D segmentation mask $\mathbf{Y}_i \in \mathbb{R}^{3\times N_{\rm class}}$ 
is provided as target value, and a distinct latent vector $\mathbf{z}_i$ is optimized along with the parameters $\theta$. To achieve the updates, the loss function $L(\hat{\mathbf{Y}}_i, \mathbf{Y}_i)$ combines a Soft Dice and a cross-entropy terms. Let $\mathbf{X}_i = \{ \mathbf{x}_i^j \}_{j=1}^M$  denote the set of $M$ 3D coordinates used to describe shape $i$, $ \mathbf{Y}_i = \{ \mathbf{y}_i^j \}_{j=1}^M$ represents the associated ground truth (GT) 
class values, and $\hat{\mathbf{Y}}_i = \{ f_\theta(\mathbf{x}_i^j, \mathbf{z}_i) \}_{j=1}^M$ corresponds to the predicted occupancy probabilities per class.
The overall minimization problem is formalized as:


\begin{equation}
\min_{\theta,\left\{\mathbf{z}_i\right\}_{i=1}^N} \frac{1}{N} \sum_{i=1}^N \left( L(\hat{\mathbf{Y}}_i, \mathbf{Y}_i) + \lambda | \mathbf{z}_i |_2^2 \right),
\end{equation}
%
%
where $\lambda$ is a regularization coefficient.
In practice, the latent vectors $\mathbf{z}_i$ are initialized randomly $\mathbf{z}_i \sim \mathcal{N}(0, 0.1^2)$, for each shape, and gradient-based optimization is used to update both the latent codes $\mathbf{z}_i$ and the network parameters $\theta$ jointly. This latent conditioning enables the use of a single shared classifier to predict the occupancy grids for the entire population $\{\hat{\mathbf{Y}}_i\}_{i=1}^N$. 


\subsection{Inference phase}


After training, we select the most informative slices (c.f. Sec.~\ref{subsec:selection_slices}) to be segmented before making an inference. 
Given the expert annotations on the selected slices, $f_\theta$ allows to predict a new unseen shape $\mathbf{\hat{Y}}_{\rm new}$ from the limited observations. In contrast to the training phase, only this subset of slices
serves to optimize the unknown latent variable $\mathbf{z}_{\rm new}$ and guide the prediction of the full 3D shape $\hat{\mathbf{Y}}_{\rm new}$. Formally, we minimize $ \mathbf{z}_{\rm new} = \arg\min_z \mathcal{L}(\hat{\mathbf{Y}}_{\rm new}, S) + \lambda \| \mathbf{z} \|_2^2$ and predict $\hat{\mathbf{Y}}_{\rm new} = \{ f_\theta(\mathbf{x}_{\rm new}^j, \mathbf{z}_{\rm new}) \}_{j=1}^M$ for all (or a subset of) coordinates $\in \{\mathbf{x}_{\rm new}^j\}_{j=1}^M$.

\subsection{Selection of Informative Shape Slices} 
\label{subsec:selection_slices}

Amiranashvili et al. \cite{AMIRANASHVILI2024103099} assumes a regular spacing between the slices to segment.
Our method proposes instead to take into account the specificity of anatomical shapes and the medical physicians annotations preferences. The goal is to reduce to the minimum the amount of necessary manual segmentations. To this end, we determine a minimal subset of slices starting from the middle slice for each shape. The next step of the algorithm consists in interactively adding new slices by evaluating the trained model and selecting the slices where the predictions made the more errors.
The final output is a list of slices that will be stored and used for inference on the test set. 


\subsubsection{Use case 1 (UC1): Organs at risk segmentation for brain cancer patients.}

\begin{figure}[htbp!]
\centering
\includegraphics[width=\textwidth]{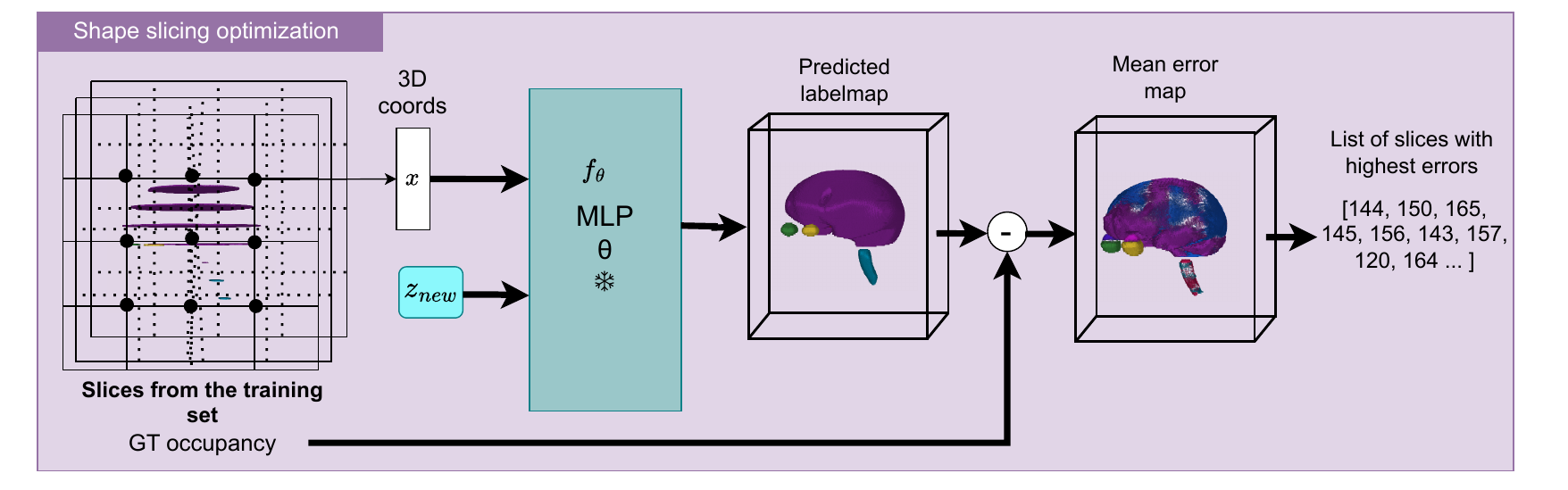}
\caption{Overview of error map generation and slice selection for UC1} 
\label{fig:diagram_usecase1}
\end{figure}

This use case explores multi-organ segmentation within the head.
In this use case, after registration, the positions of the organs do not vary a lot from one patient to another. Since the goal is to minimize the annotations for future patients, the slice selection is performed on the training set, for which we have the full volume available. The minimal subset of slices is the middle slice of each organ in the axial direction.
An initial set of inferred volumes are computed from this subset. Each prediction is then subtracted to its GT to obtain a mean error map (see Fig~\ref{fig:diagram_usecase1}). The slice index that obtains the highest error in this map is retained for posterior inferences on the test set. Iteratively, this phase is repeated until a predetermined maximum number of slices is reached.
\subsubsection{Use case 2 (UC2) : Few-Shot Annotation of a New Database with Domain Shift}
\begin{figure}[H]
\centering
\includegraphics[width=\textwidth]{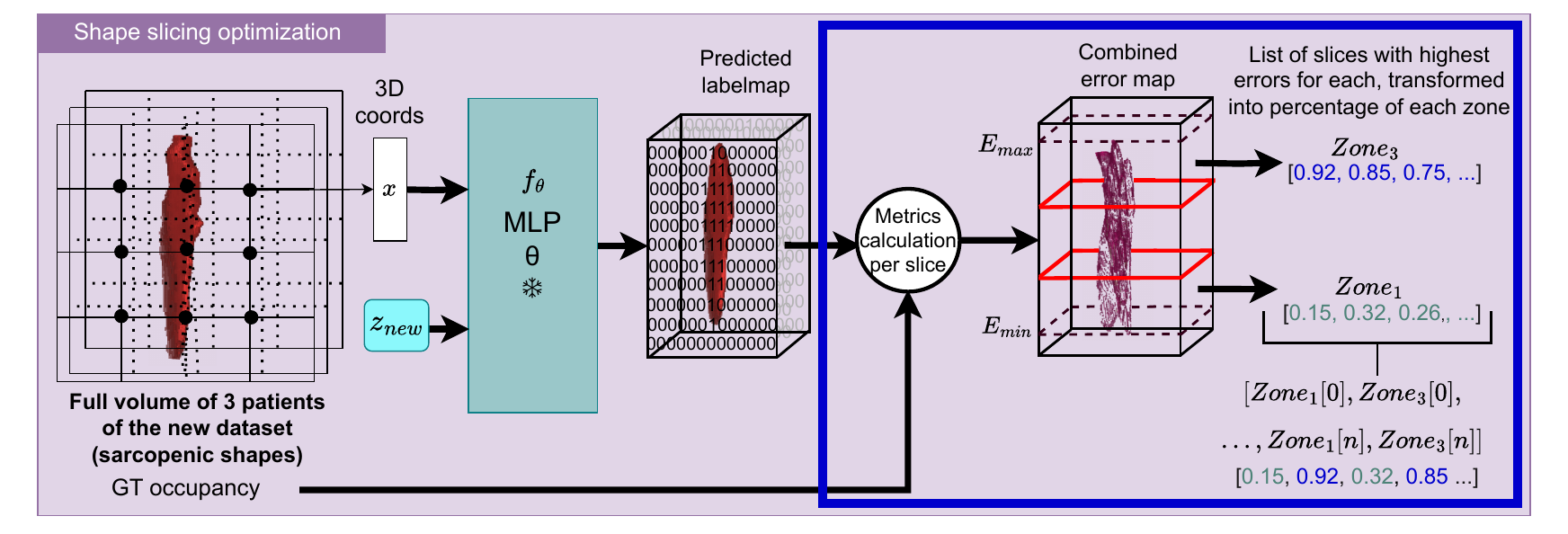}\\
\vspace{0.5em}
\includegraphics[width=\textwidth]{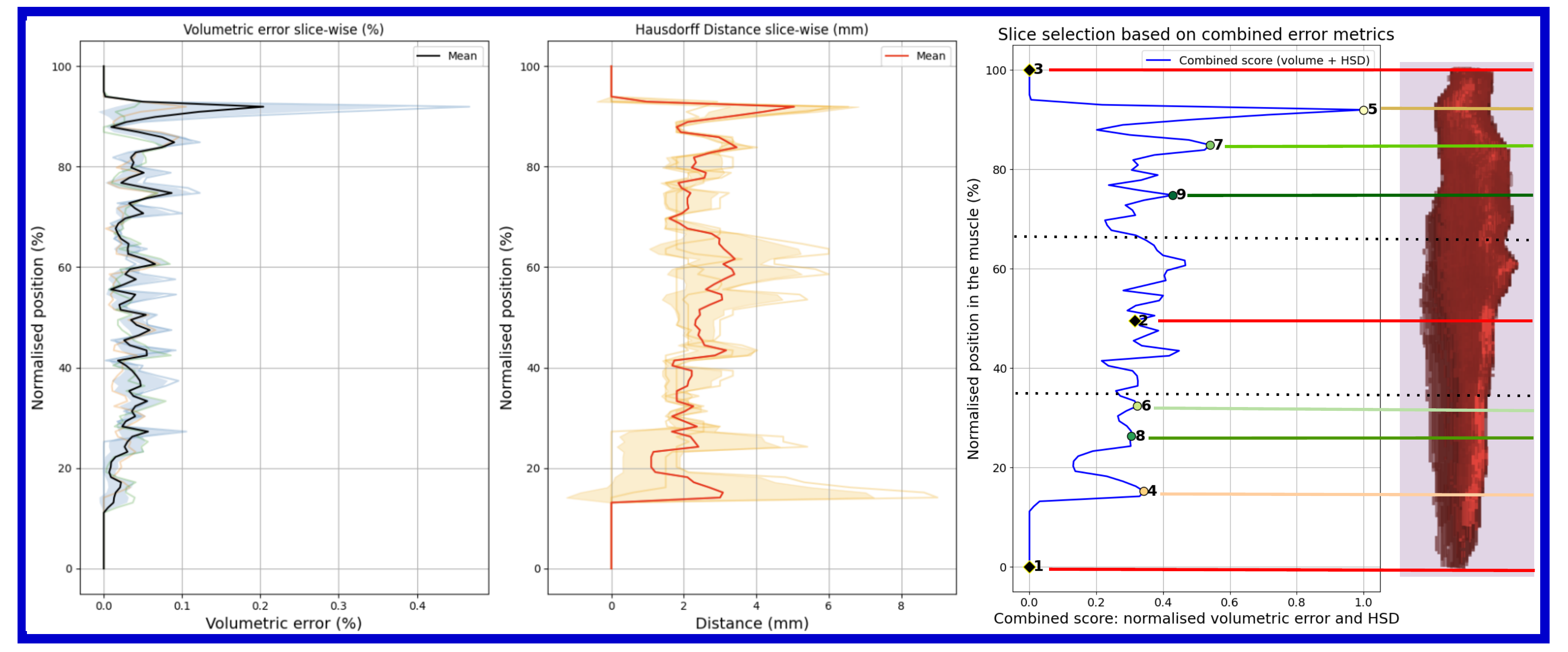}
\caption{Overview of combined error map generation and slice selection for UC2 (\textbf{top}) and detailed metric calculation and slice selection (\textbf{bottom}).}
\label{fig:shape_slicing_DIASEM}
\end{figure}

This use case targets the creation of a new database involving `abnormal' muscle shapes (e.g., sarcopenia), adapting the shape prior from a model trained on non-sarcopenic subjects. As shown in Fig.~\ref{fig:shape_slicing_DIASEM}, three fully segmented labelmaps from this new population are used during the adaption phase to evaluate the model under domain shift.
To capture segmentation performance, we compute the Hausdorff Distance and volumetric errors slice-wise after normalizing muscle lengths from 0\% (distal insertion) to 100\% (proximal insertion) across subjects. This normalization allows interpolation and averaging of metrics per relative slice position (and tackles inter-subject variability in muscle length). Example curves for both metrics appear on the lower left and center of Fig.~\ref{fig:shape_slicing_DIASEM}. To obtain a single score per relative slice we normalize each metric to a [0, 1] range and average both. The resultant curve is shown in the lower-right of Fig.~\ref{fig:shape_slicing_DIASEM}.
A key challenge in building new muscle databases is the variability in image dimensions and muscle lengths across individuals. Therefore, the minimum slice subset in UC2 includes the muscle insertions (first and last visible slices) and the mid-axial slice. Then, the slice selection focuses on regions with the highest errors within the regions characterized by the most pronounced variations, namely the distal zone (zone 1) and the proximal zone (zone 3). These zones are obtained by splitting the muscle in three according to its length. The final slice list, expressed as a percentage of muscle length, alternates between zones 1 and 3 after the three insertion/midsection slices (see Fig~ \ref{fig:shape_slicing_DIASEM}) with a constraint of at least 5 slices from previous selected slices. The bottom-right graph of the figure depicts this selection process: the first three slices (in red), are selected initially, followed by an iterative addition of slices indexed from 4 to 9. This approach enables a controlled comparison of the predictions with the baseline. 

\section{Experiments}
\label{sec:experiments}




\textbf{Implementation Details}
Most of the model parameters are kept as in \cite{AMIRANASHVILI2024103099} including the MLP architecture: the learning rate was set to  $1.0e^{-4}$ and to $1.0e^{-3}$ for the AD. The MLP has 8 layers, the output of the 4th layer is concatenated with the 3D coordinates, and the dimension of $\mathbf{z}$ is $128\cdot N_{\rm class}$. The dimensions of the layers were adapted to fit the input dimensions (3D coordinates concatenate with $\mathbf{z}$). 
The model was updated using ADAM during 2500 epochs for training on two GPU NVIDIA GeForce RTX (4000 for UC1 and A6000 for UC2). The slices are provided in an axial plane, for both training and inference. The baseline for this experiment is Amiranashvili et al \cite{AMIRANASHVILI2024103099} with the same number of slices. This baseline is referred to as strategy 0, and our method is strategy 1. 
Other details specific to each use case are further developped in the next sections. 



\subsection{UC1 : Organs at Risk Segmentation}

The number of inference epochs in this case is 300, which is sufficient to reach convergence. This result is achieved within 210 seconds, so approximately 3 minutes and a half for the entire volume. During training, only one slice out of two were used due to memory constraints.


\subsubsection{Dataset} 

The dataset consists of 20 brain {CT} images with segmentation masks, split into 8 patients for training, 2 for validation, and 10 for testing.
To ensure spatial consistency across the dataset, all patients were rigidly registered to a common reference space.  The same organs are present in all patients. Their relative volume are 0.06\textperthousand \ for the Optic Chiasma (OC), 0.8\textperthousand \ for the left (LE)  and right eye (RE), 0.5 \textperthousand \ for the spinal cord (SC) and 134 \textperthousand \ for the brain (B).

\begin{figure}[!htbp]
\centering
\includegraphics[width=0.7\textwidth]{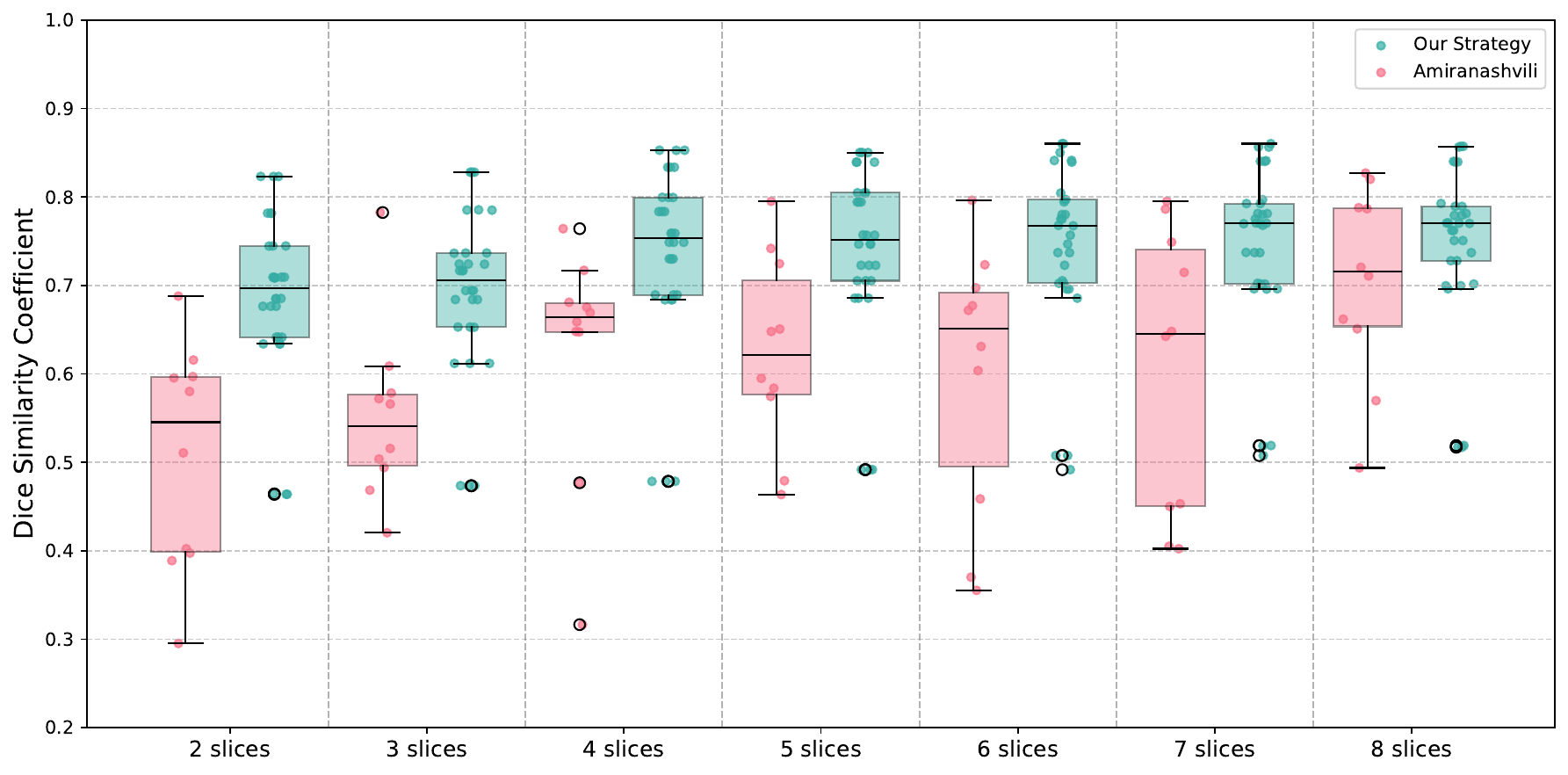}
\includegraphics[width=0.7\textwidth]{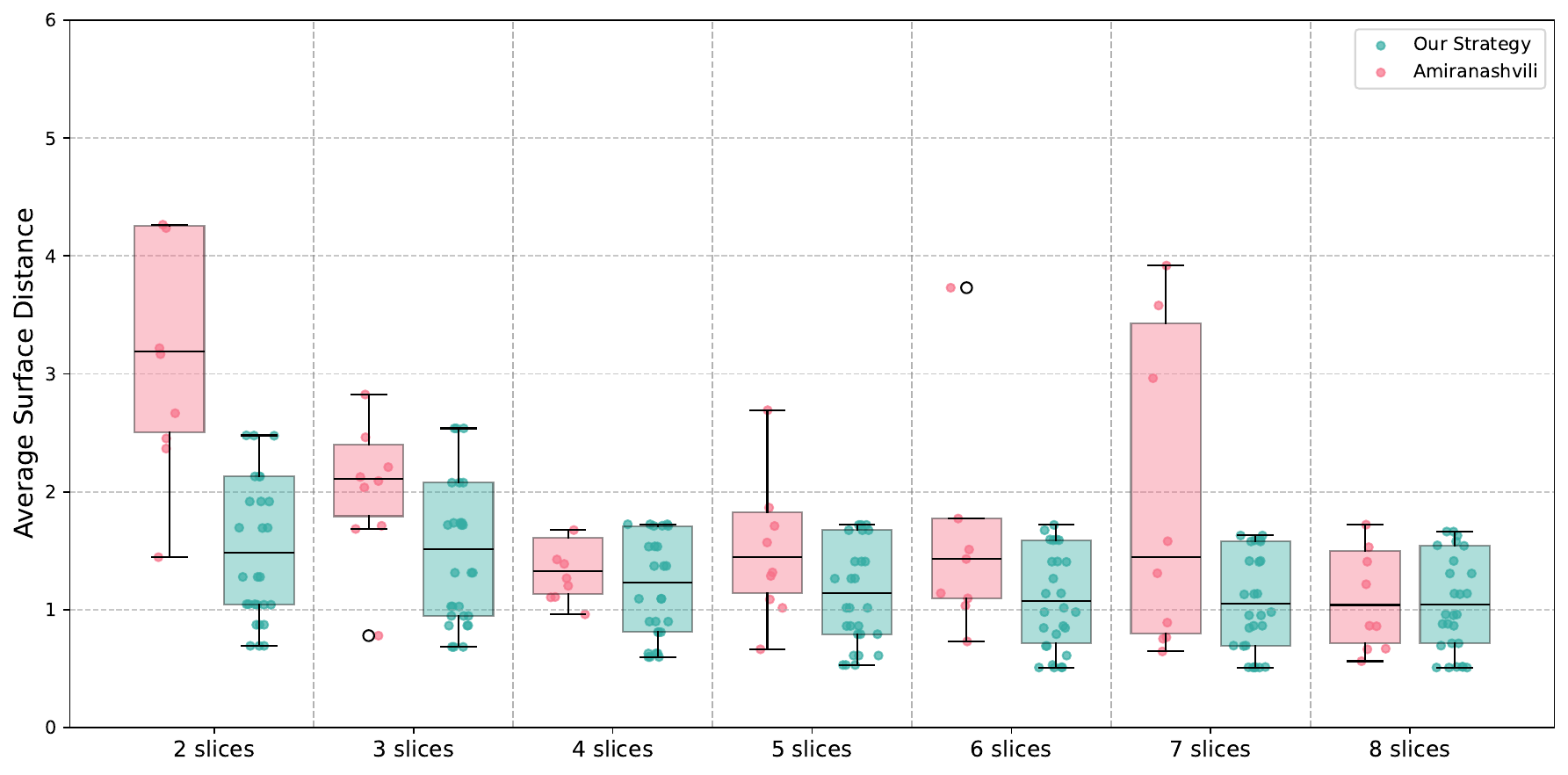}
\vspace{0.5em}
\includegraphics[width=0.7\textwidth]{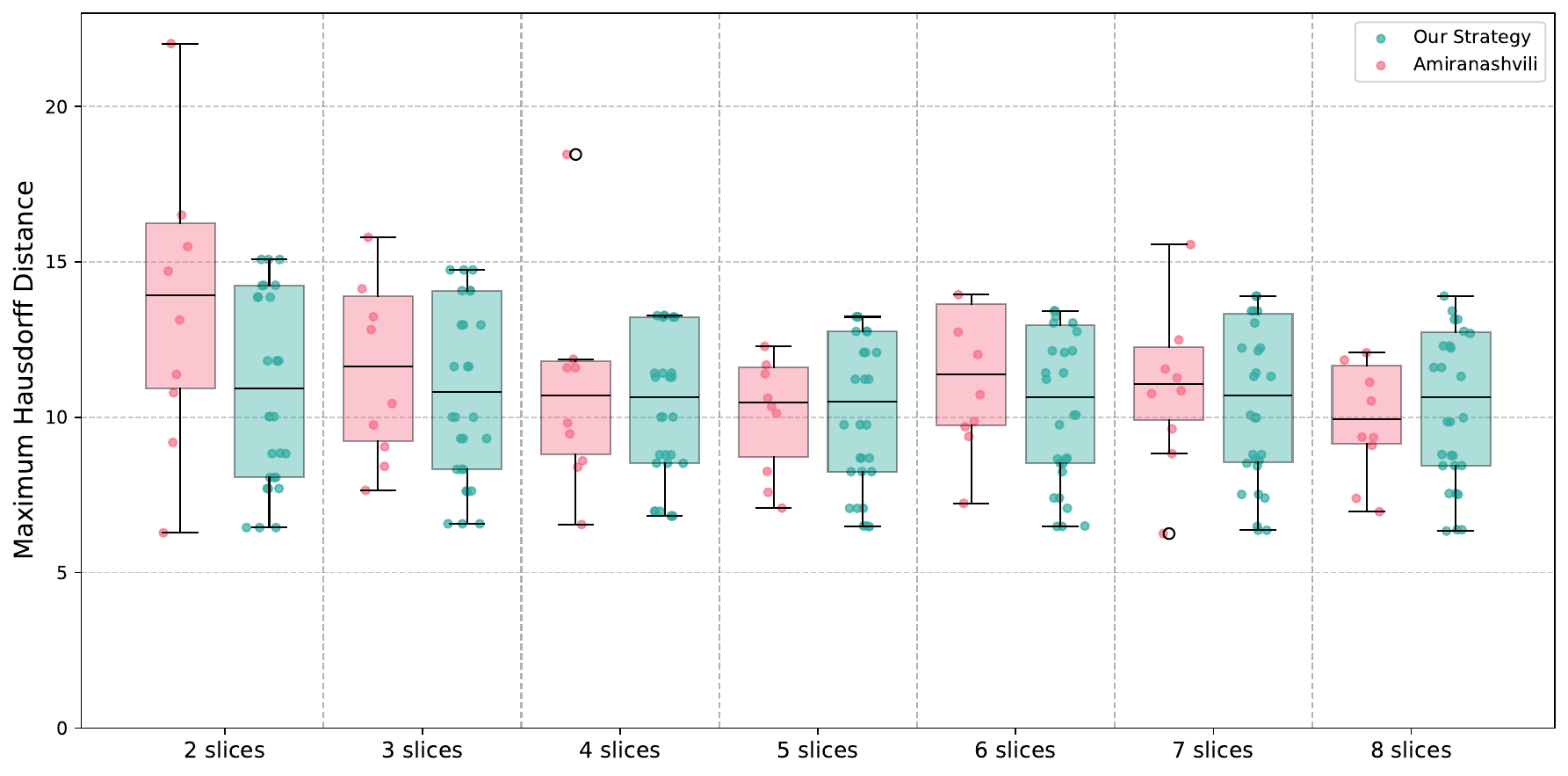}
\caption{Comparison across different slice numbers on the test set.}
\label{fig:quantitative_perf_CEMMTAUR}
\end{figure}


\subsubsection{Quantitative Evaluation}

Fig.~\ref{fig:quantitative_perf_CEMMTAUR} shows the DSC, ASD, and the maximum Hausdorff distance for different numbers of slices with both strategies. Our slicing strategy outperforms the baseline on every metric when inferring with two or three slices. In particular, the difference is the highest when using two slices: 0.70 DSC for our method vs 0.55 for the baseline. The boxplot quartile range is much
more spread for strategy 0. The ASD shows the same trend: 3.2 vs 1.5.
For the maximum Hausdorff distance, the average is 14 for two slices going up to 23 for the baseline, while the average for strategy 1 is 12, with a maximum of 15.
Furthermore, for whichever amount of slices, our method's average DSC is always better, and the interquartile range remains around 0.2, regardless of slice count. In contrast, the DSC from the baseline method has a boxplot range of up to 0.4, even when using 7 slices. The same pattern is seen in the ASD, with wider value ranges (up to 3) and more outliers.


\begin{table}
\caption{Average DSC $\pm$ standard deviation according to the number of slices for each organ : the Optic Chiasma (OC), the left and right eye (LE, RE), the spinal cord (SC) and the brain (B). Strategy (S) 0: baseline,  1: ours. Best results base in bold.} \label{tab:avg_dice_per_organ}

\begin{tabular}{llccccccc}
\toprule
& S  & 2 sl. & 3 sl. & 4 sl. & 5 sl. & 6 sl. & 7 sl. & 8 sl. \\
\midrule
OC & 0 & 0.98 $\pm$ 0.0 & \textbf{0.99} $\pm$ 0.0 & \textbf{0.99} $\pm$ 0.0 & \textbf{0.99} $\pm$ 0.0 & \textbf{0.99} $\pm$ 0.0 & \textbf{0.99} $\pm$ 0.0 & \textbf{0.99} $\pm$ 0.0 \\
   & 1 & \textbf{0.99} $\pm$ 0.0 & \textbf{0.99} $\pm$ 0.0 & \textbf{0.99} $\pm$ 0.0 & \textbf{0.99} $\pm$ 0.0 & \textbf{0.99} $\pm$ 0.0 & \textbf{0.99} $\pm$ 0.0 & \textbf{0.99} $\pm$ 0.0 \\
\midrule
LE & 0 & 0.12 $\pm$ 0.2 & 0.08 $\pm$ 0.1 & 0.11 $\pm$ 0.1 & 0.20 $\pm$ 0.2 & 0.13 $\pm$ 0.2 & 0.28 $\pm$ 0.1 & 0.32 $\pm$ 0.2 \\
   & 1 & \textbf{0.44} $\pm$ 0.1 & \textbf{0.41} $\pm$ 0.2 & \textbf{0.43} $\pm$ 0.1 & \textbf{0.44} $\pm$ 0.1 & \textbf{0.41} $\pm$ 0.2 & \textbf{0.41} $\pm$ 0.2 & \textbf{0.43} $\pm$ 0.2 \\
\midrule
RE & 0 & \textbf{0.54} $\pm$ 0.2 & \textbf{0.33} $\pm$ 0.2 & 0.51 $\pm$ 0.2 & \textbf{0.55} $\pm$ 0.2 & \textbf{0.67} $\pm$ 0.1 & 0.61 $\pm$ 0.2 & 0.58 $\pm$ 0.2 \\
   & 1 & 0.24 $\pm$ 0.2 & 0.27 $\pm$ 0.2 & \textbf{0.55} $\pm$ 0.2 & 0.54 $\pm$ 0.2 & 0.62 $\pm$ 0.1 & \textbf{0.62} $\pm$ 0.1 & \textbf{0.62} $\pm$ 0.1 \\
\midrule
SC & 0 & 0.35 $\pm$ 0.2 & 0.43 $\pm$ 0.2 & 0.45 $\pm$ 0.2 & 0.48 $\pm$ 0.3 & 0.57 $\pm$ 0.3 & 0.57 $\pm$ 0.3 & 0.63 $\pm$ 0.3 \\
   & 1 & \textbf{0.76} $\pm$ 0.3 & \textbf{0.77} $\pm$ 0.3 & \textbf{0.78} $\pm$ 0.3 & \textbf{0.78} $\pm$ 0.3 & \textbf{0.77} $\pm$ 0.3 & \textbf{0.77} $\pm$ 0.3 & \textbf{0.77} $\pm$ 0.3 \\
\midrule
B  & 0 & 0.36 $\pm$ 0.2 & 0.41 $\pm$ 0.2 & 0.45 $\pm$ 0.3 & 0.47 $\pm$ 0.3 & 0.57 $\pm$ 0.3 & 0.60 $\pm$ 0.3 & 0.69 $\pm$ 0.3 \\
   & 1 & \textbf{0.77} $\pm$ 0.3 & \textbf{0.78} $\pm$ 0.3 & \textbf{0.76} $\pm$ 0.3 & \textbf{0.77} $\pm$ 0.3 & \textbf{0.75} $\pm$ 0.3 & \textbf{0.75} $\pm$ 0.3 & \textbf{0.75} $\pm$ 0.3 \\
\bottomrule
\end{tabular}

\end{table}


Table \ref{tab:avg_dice_per_organ} shows the average DSC obtained by strategies 0 and 1 for each organ, sorted by increasing size. For the smallest organ, the optic chiasma, the dice score varies very little between strategies, which could be explained by the inter-patient variation.
The eyes show more variable performances when using fewer than 6 slices. 
The benefits of our algorithm are more evident for larger organs: the spinal cord and the brain show a dice score improvements of +0.41 with two slices, and +0.34/0.37 with three slices. The gap between the two methods gradually decreases when using more slices, but our strategy consistently outperforms the baseline for these organs. Furthermore, our strategy outperforms the baseline for all organs when using more than six slices.


\subsubsection{Qualitative Evaluation}

\begin{figure}
\centering
\includegraphics[width=0.49\textwidth]{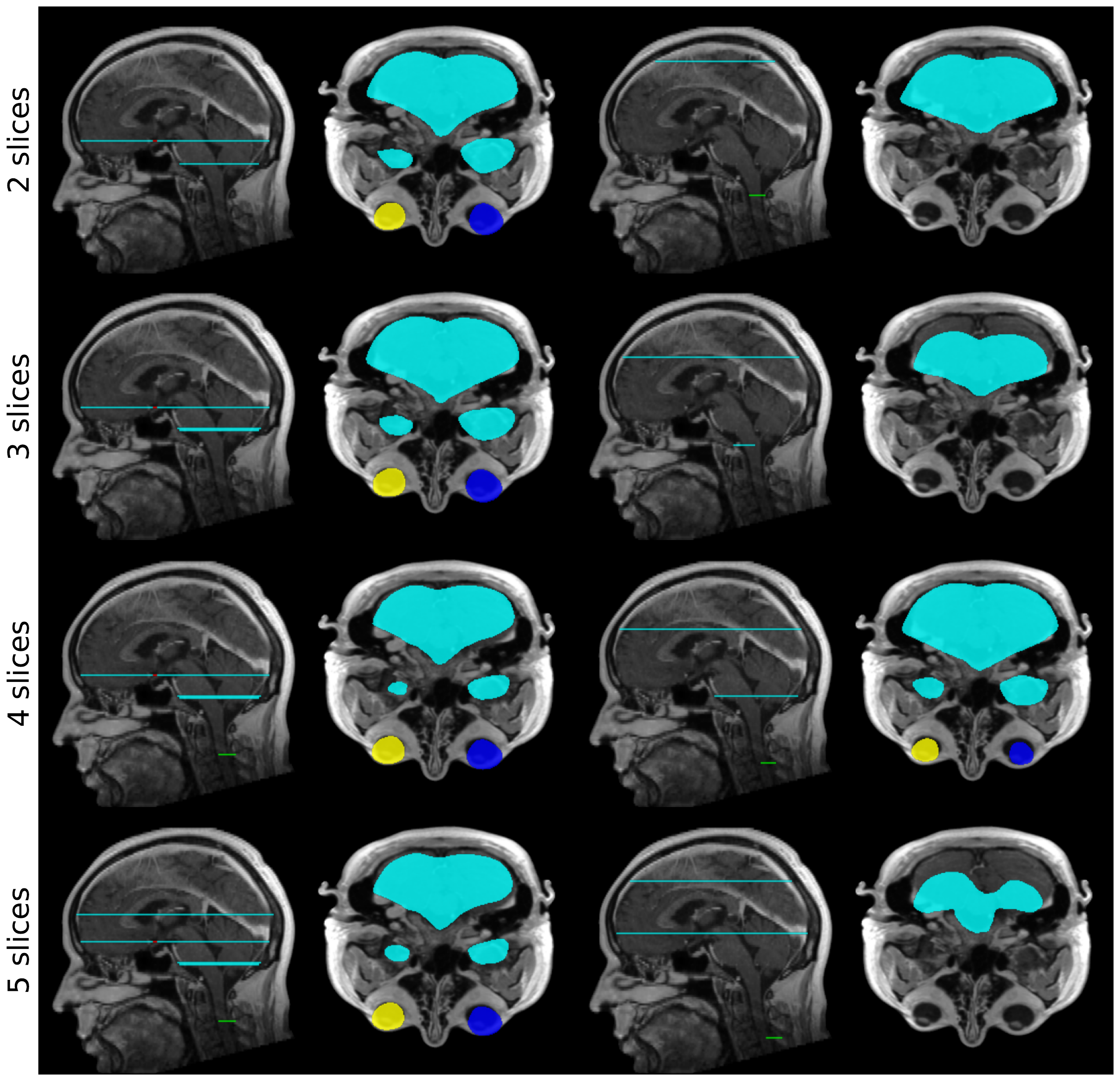}
\includegraphics[width=0.49\textwidth]{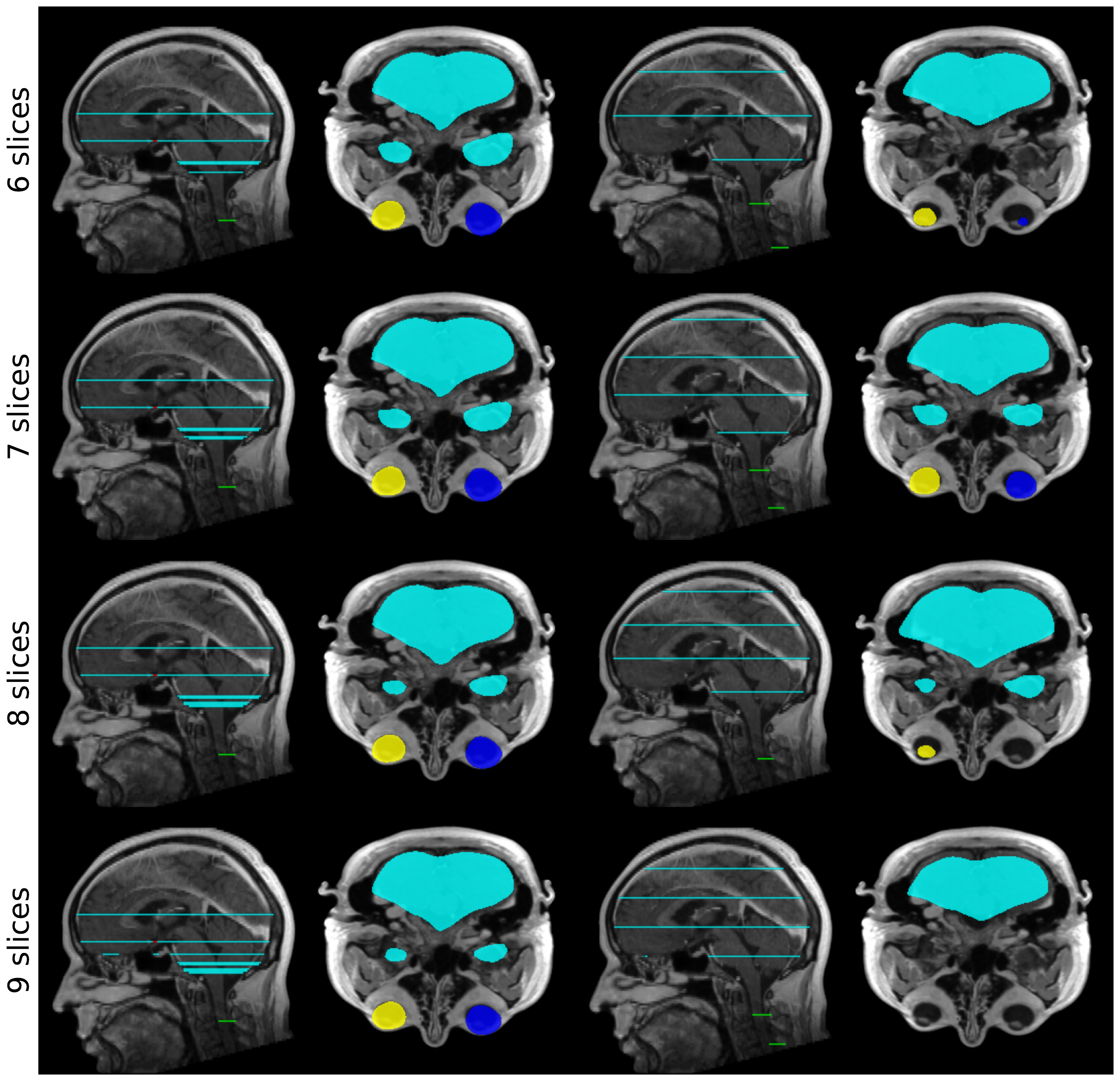}
\caption{Slices and predictions used for each amount of slices. Cyan is the brain, dark blue is the left eye, yellow is the right eye and green is the spinal cord.} 
\label{fig:slices_selected_CEMMTAUR}
\end{figure}

This section presents the results on a test patient. The segmented shapes are displayed on the associated MRI for context only, the image information is not used for training nor inference. 
The first and third columns of Fig.~\ref{fig:slices_selected_CEMMTAUR} show the slices selected by our method and the baseline, respectively. The second and fourth columns display an axial view of the volume predictions from inferring with subset of slices in each case. 
These images illustrate that our algorithm sometimes selects slices that are very close together. This explains why the performance does not improve significantly beyond four or five slices. The predictions are consistent with the quantitative results: our method produces more robust segmentations, particularly when using less than four slices. Notably, the eye segmentation is inconsistent for strategy 0: in some cases (e.g., eight slices), only one eye is segmented, while in others (e.g., two, three, five, or even nine slices), neither eye appears.


\subsubsection{Discussion}

The proposed interactive method is more robust
and achieves better scores than the baseline across all metrics and qualitatively.
Possible improvements include optimizing the initial slice selection in a single step. Since organs like the eyes and brain, or the brain and optic chiasma, have overlapping regions due to differences in organ size, it may be sufficient to slice through the center of the eyes or optic chiasma to capture part of the brain as well.



\subsection{UC2: Few-Shot Annotation of a Database with Domain Shift}

Once the slice IDs selected on the 3 few-shot volumes according to Sec.~\ref{subsec:selection_slices}-UC2, multiple test inferences on the new domain were performed based on the imposed number of slices, ranging from 3 to 9 slices. Each inference was run for 1500 epochs on a GPU NVIDIA GeForce RTX 3090 for all test subjects, taking approximately 30 seconds per subject.

\subsubsection{Datasets} Two in-house datasets were used for this use case. 
The \textit{Young subjects dataset} comprised 3 acquisitions, with 3 different compression of the probe: standard and minimal compression and using a gel pad of 3DUS. 45 images of 15 healthy participants, 8 males, aged $27 \pm 2$ years (height: $172 \pm 6$ cm, weight: $63 \pm 6 $kg), were acquired and manually segmented by an expert clinician.
4 muscles were included in the initial study \cite{HUET2024}, but we focus on the right Rectus Femoris (RF). 
The \textit{DIASEM dataset} is composed of 3DUS scans of 58 ``older adults" (age $\geq 75$), hospitalized in rehabilitation or geriatric medicine, with 23 diagnosed with sarcopenia. The protocol from \cite{huet2024_journal} was used to acquire the data. The same expert as for \textit{Young subjects dataset} created manual segmentations of three muscles per patient; as earlier, we focus on the right RF.
The labelmaps of both datasets were registered using the RF/muscle barycenter. As in \cite{piecuch2025unsupervisedanomalydetectionimplicit},  ``normal''  dataset were created and composed of all  volumes in the \textit{Young subjects dataset } and the non-sarcopenic older subjects of the DIASEM dataset. $77\%$ of normal data are used during the train. The test set here, is only composed of \textit{DIASEM}. Only the sarcopenic subset were used for the adaption and inference.




\subsubsection{Quantitative Evaluation}
\begin{figure}
\centering
\includegraphics[width=0.7\textwidth]{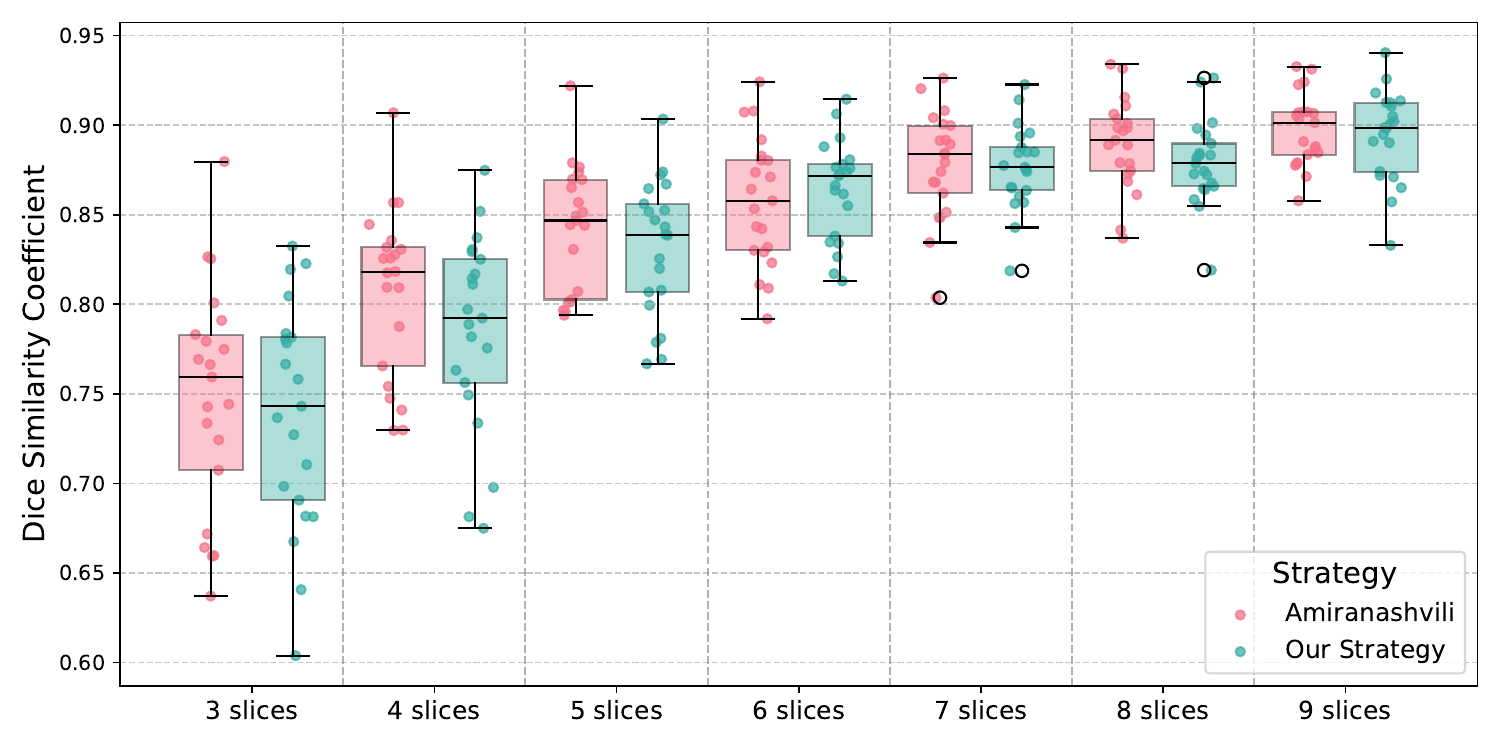}
\includegraphics[width=0.7\textwidth]{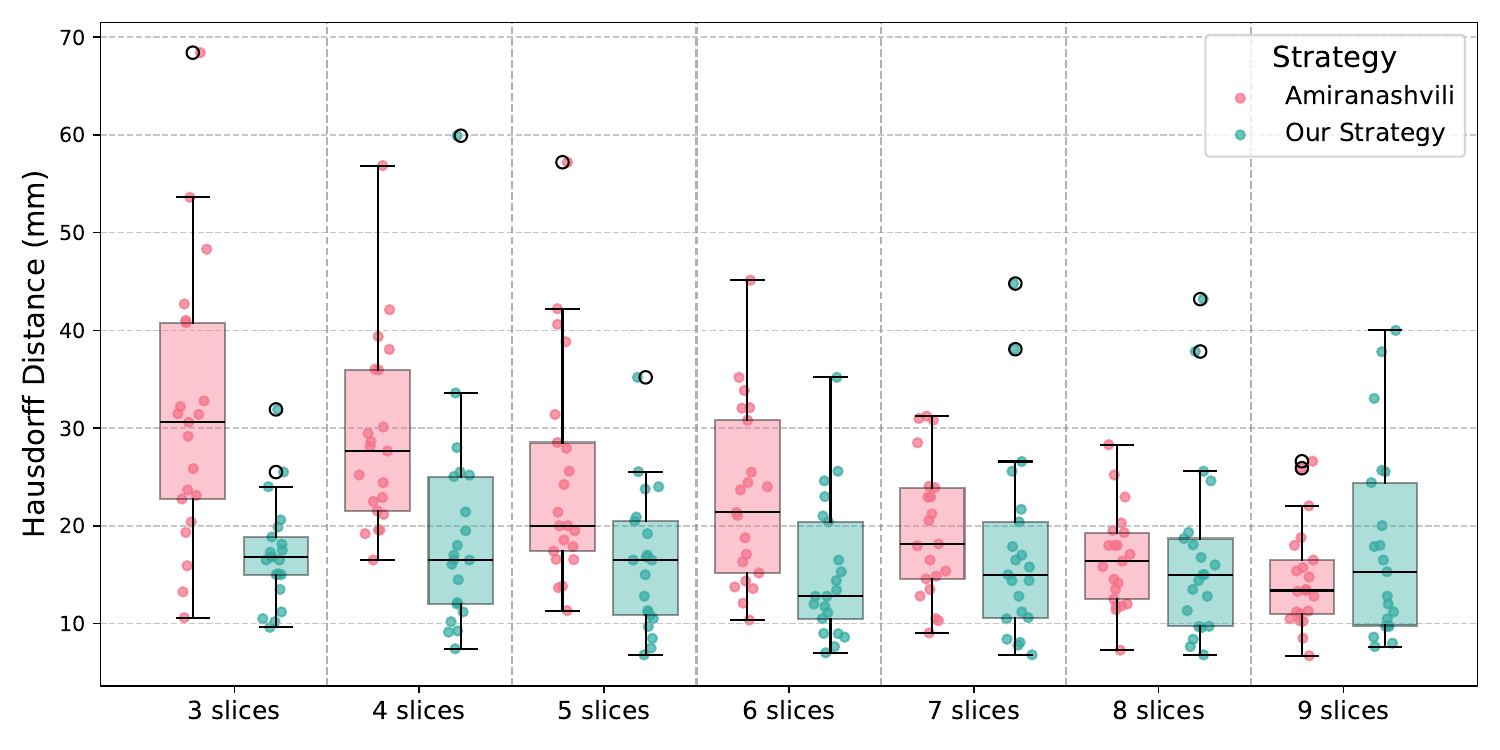}
\includegraphics[width=0.7\textwidth]{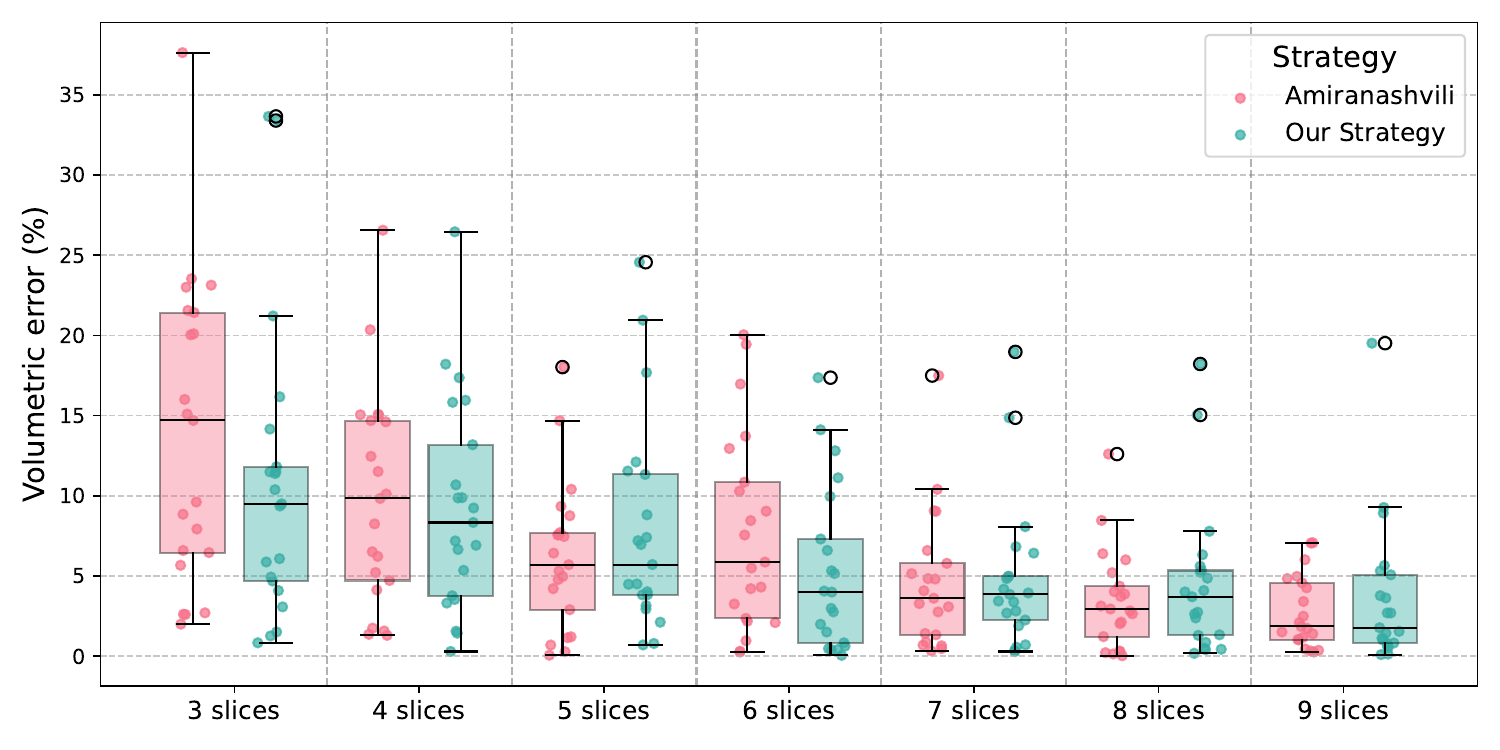}
\caption{Comparison across different slice numbers on the test set} 
\label{fig:quantitative_perf_DIASEM}
\end{figure}

The quantitative results for UC2 are presented in Fig~\ref{fig:quantitative_perf_DIASEM}, focusing on three metrics: DSC, HSD, and volumetric error (in percent). While the DSC values are slightly lower than those obtained using the baseline (except for the six slices configuration) our method demonstrates superior performance on the other two metrics, even when fewer slices are used. Specifically, the average Hausdorff distances achieved by our method remain consistently around 15mm across all slice configurations. In contrast, the baseline remains above 15mm up to eight slices and exceeding 20mm between three and six slices, with a maximum of 30mm observed for the lowest slice count. Furthermore, our method exhibits lower standard deviations in HSD up to six slices, highlighting its robustness when operating with limited input, an essential property for accelerating segmentation labelmap generation for novel anatomical shapes. Regarding the volumetric error, our approach also yields lower average errors from three to six slices and at nine slices. Notably, it maintains an error around 10\% for three and four slices and drops below 5\% starting from six slices. In comparison, the baseline reports higher errors, ranging from 15\% to 10\% for the same low-slice settings.
Similarly, the boxplots are more compact, demonstrating improved robustness from the outset. 


\subsubsection{Qualitative Evaluation}

\begin{figure}
\centering
\includegraphics[width=0.8\textwidth]{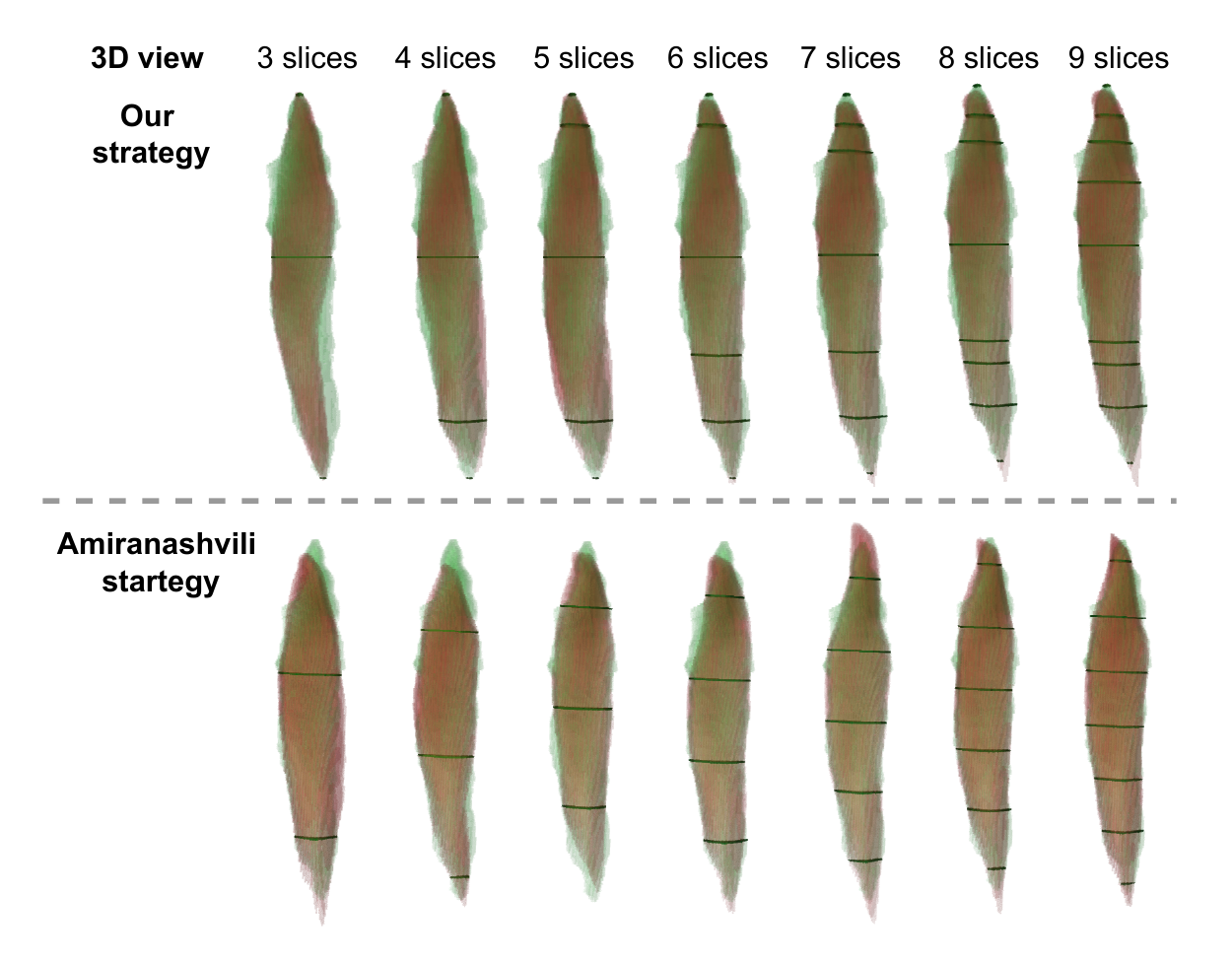}
\caption{Comparison of qualitative results on different number of slices of Amiranashvili strategy (\textbf{bottom}) with ours (\textbf{top}) on UC2. Supperpostion of prediction in \textbf{red}, with the GT slices inputed to the model in \textbf{dark green} and the GT shape in \textbf{lightgreen}.} 
\label{fig:qualitativ_perf_DIASEM}
\end{figure}

Fig. \ref{fig:qualitativ_perf_DIASEM} compares qualitative results from the baseline (bottom) and our method (top), across varying numbers of input slices. Input GT slices (dark green) are overlaid on GT volumes (light green), with predictions in red. The slice count in the baseline may differ from the indicated number, as it selects slices across the full labelmap, sometimes outside the actual segmentation area (i.e., background). In contrast, our strategy constrains the slice selection directly within the object we aim to segment. This highlights that 
our strategy aligns better with clinical practice, allowing experts to annotate directly within the region of interest.
Also, by ensuring the inclusion of muscle insertions and the central slice, our method reaches good performance with fewer slices (e.g., three and four), outperforming the baseline early on (see Fig. \ref{fig:qualitativ_perf_DIASEM}). As slice count increases (e.g., seven to nine), the baseline improves, due to better longitudinal coverage of elongated structures like the RF, though it misses critical areas such as insertions. Our zone-based selection offers both full coverage and focus on challenging regions.

\subsubsection{Discussion}
In UC2, which involves a domain shift (e.g., sarcopenic patients), our method designed to prioritize clinically relevant regions and accelerate manual segmentation, demonstrates several advantages when only a few slices are provided. It outperforms the baseline with only three and four slices, particularly in terms of Hausdorff Distance and volumetric error.
Qualitatively, our anatomically guided slice selection (e.g., insertions and center) avoids background regions and supports realistic clinical workflows, unlike Amiranashvili’s approach. This targeted strategy enables stronger performance from the start and is better suited to practical, time-efficient segmentation.



\section{Conclusion}


This paper proposed an interactive method to select a subset of informative slices to accelerate manual segmentation on new data based on a learned shape prior.
The application of the method in two practical medical use cases demonstrated an improvement compared to the initial approach proposed by Amiranashvili et al. \cite{AMIRANASHVILI2024103099}. The selection of specific slices allows for better reconstruction, particularly when the number of slices is very limited. 
In the more complex UC2 with domain shift, experiments reveal certain limitations, especially for long anatomical structures, which show more mixed performance as the number of slices increases. However, our method aligns better with the clinical practice.

An interesting perspective is to leverage the probability maps produced by the model to identify regions of high uncertainty and selectively add slices in those areas. This could provide complementary information to guide the annotation or refinement process more effectively. For the initialization step, rather than defaulting to the middle slice of each organ, a more strategic approach would be to choose slices that include the highest number of segmented organs. This would maximize the anatomical information available from the start and potentially improve the overall segmentation performance. An extension to other muscles, or even a configuration with multiple muscles at once, could be considered for UC2 to increase the generability of this study.

\subsubsection{Compliance with ethical standarts}
\label{sec:ethic}
This protocol was approved by the local ethics committee of the Nantes University Hospital "Groupe Nantais d'éthique dans le Domaine de la Santé" : AVIS 24-6-01-191. 

\subsubsection{Acknowledgements}
This work was partially supported by the CEMMTAUR project. The authors would like to thank the University Hospitals of Rennes and Nantes for their involvement in data acquisition and the creation of segmentation label maps. We also thank the clinical collaborators for their valuable insights, which helped guide the development of the method.

%
%
%
\bibliographystyle{splncs04}
%

\bibliography{biblio-macros,biblio}
\end{document}